\documentclass{article}

     \usepackage[preprint]{neurips_2025}

\usepackage[utf8]{inputenc} 
\usepackage[T1]{fontenc}    
\usepackage{hyperref}       
\usepackage{url}            
\usepackage{booktabs}       
\usepackage{amsfonts}       
\usepackage{nicefrac}       
\usepackage{microtype}      
\usepackage{xcolor}         

\usepackage{graphicx}
\usepackage{amsmath}
\usepackage{subcaption}
\usepackage{caption}
\usepackage{booktabs}
\usepackage{cleveref}
\usepackage{dirtytalk}
\usepackage{verbatim}

\title{Compressed Computation: Dense Circuits in a Toy Model of the Universal-AND Problem}

\author{%
  Adam Newgas \\
  Independent\\
  \texttt{adam2020@newgas.net} \\
}

\begin{document}
\maketitle

\begin{abstract}
Neural networks are capable of superposition - representing more features than there are dimensions. Recent work considers the analogous concept for computation instead of storage, proposing theoretical constructions. But there has been little investigation into whether these circuits can be learned in practice.

In this work, we investigate a toy model for the Universal-AND problem which computes the AND of all $m\choose 2$ pairs of $m$ sparse inputs. The hidden dimension that determines the number of non-linear activations is restricted to pressure the model to find a compute-efficient circuit, called compressed computation.

We find that the training process finds a simple solution that does not correspond to theoretical constructions. It is fully dense - every neuron contributes to every output. The solution circuit naturally scales with dimension, trading off error rates for neuron efficiency. It is similarly robust to changes in sparsity and other key parameters, and extends naturally to other boolean operations and boolean circuits. We explain the found solution in detail and compute why it is more efficient than the theoretical constructions at low sparsity.

Our findings shed light on the types of circuits that models like to form and the flexibility of the superposition representation. This contributes to a broader understanding of network circuitry and interpretability.
\end{abstract}





\begin{figure}[htbp]
  \centering
  \begin{minipage}[c]{0.44\textwidth}
    \centering
    \includegraphics[width=\linewidth]{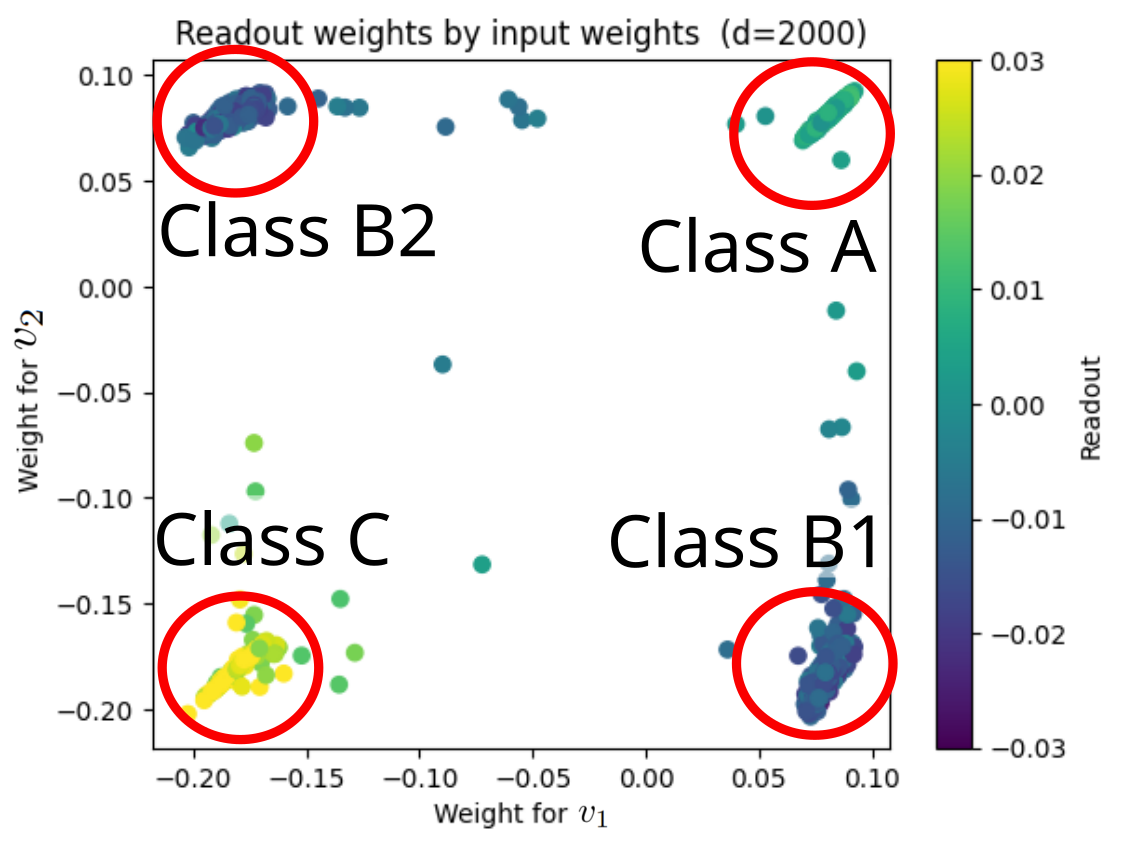}
  \end{minipage}
  \hfill
  \begin{minipage}[c]{0.50\textwidth}
    \centering
\[\begin{aligned} 
\text{Class A}  & \quad & y_i = \operatorname{ReLU}(&uv_1+uv_2+X_i+b) \\
\text{Class B1} & \quad & y_i = \operatorname{ReLU}(&uv_1+lv_2+X_i+b) \\ 
\text{Class B2} & \quad & y_i = \operatorname{ReLU}(&lv_1+uv_2+X_i+b) \\
\text{Class C}  & \quad & y_i = \operatorname{ReLU}(&lv_1+lv_2+X_i+b)\\ \end{aligned}\]

    \resizebox{\linewidth}{!}{
      \begin{tabular}{cc|cccc|c}
        \toprule
        $v_1$ & $v_2$ & A & B1 & B2 & C & $4(A + C - B1 - B2)$ \\
        \midrule
        0 & 0 & 0.05 & 0.05 & 0.05 & 0.05 & 0 \\
        0 & 1 & 0.15 & 0    & 0.15 & 0    & 0 \\
        1 & 0 & 0.15 & 0.15 & 0    & 0    & 0 \\
        1 & 1 & 0.25 & 0    & 0    & 0    & 1 \\
        \bottomrule
      \end{tabular}
    }
      Approximate truth table for each neuron class
  \end{minipage}
  \caption{Our found circuit: For every pair of inputs (e.g. $v_1$, $v_2$) the model neurons separate into 4 classes based on response to those inputs. A linear combination of those classes recreates the AND operator, with some error (not shown).}
  \label{fig:hero}
\end{figure}

\newpage

\section{Introduction}


Models have been found to learn to store a set of sparse features $m$ in a vector of dimension $d$, where $d \ll m$. They achieve this with a linear representation of nearly orthogonal vectors per feature, called \textbf{superposition}~\citep{elhage2022toy}. By relying on the sparsity of feature activation, and tolerating a small amount of error due to overlap, an exponential number of features can effectively stored in the vector. An understanding of superposition has been critical for mechanistic interpretability of models: it has led to foundational concepts in interpretability, the development of interpretability tools (e.g. \citep{cunningham2023sparse}, \citep{ameisen2025circuit}) and provided evidence for the Linear Representation Hypothesis~\citep{park2023lrh}.

But storage of features isn't a full descriptor of how models work. Networks must also do efficient, useful computation with features. This presents two difficulties for the model. Firstly, models must be able to efficiently work with features that are already represented in superposition. This is called \textbf{computation in superposition}~\citep{hanni2024}. Secondly, models need to deal with the fact that they are limited to a few parameters and calculation units (non-linear activation of neurons), called \textbf{compressed computation}~\citep{apd}.

Just as an understanding of superposition is necessary to analyze activation space effectively, an understanding of computation will be necessary to analyze weights and circuits. Without these understandings, activations/circuits can appear inscrutably commingled. A good theory of computation will allow tools and analysis of weights and circuits in a network.

Theoretical models have been proposed that explore computation in superposition (\citep{hanni2024}, \citep{comp_in_sup_many}, \citep{comp_in_sup_complexity}). These papers find constructions that operate directly on input/output features in superposition, and establish bounds on accuracy or model size. These constructions generally rely on \textbf{sparse weights} to ensure that neurons do get too much interference from irrelevant inputs, allowing estimates to be measured. We call a construction sparse if the parameter weights are zero or negligible with probability that approaches one.

\citet{comp_in_sup_complexity} in particular notes that \say{logical operations like pairwise AND can be computed using $O(\sqrt{m'} \log m')$ neurons and $O(m' \log^2 m')$ parameters. There is thus an exponential gap between the complexity of computing in superposition versus merely representing features, which can require as little as $O(\log m')$ neurons}.

Thus even with computation in superposition, models are very likely to face bottlenecks in compute, needing compressed computation.
Our work focuses specifically on this case. We re-use the same \textbf{Universal-AND problem} setting of \citet{hanni2024}, but eliminate the main source of computation in superposition by using monosemantic input and output. A narrow hidden dimension is used to force reuse of neurons for multiple circuits, exploiting the sparsity of the inputs, and controlling the error from unrelated inputs interfering with the calculations. 


The Universal-AND problem \citep{hanni2024} is the task of efficiently emulating a circuit that takes $m$ sparse boolean inputs and produces $m \choose 2$ outputs that each compute the AND operation of a given pair of inputs, described further in \cref{background}. We train toy models with one layer of ReLU on this problem for various settings of sparsity, $s$, and hidden dimension size, $d$.



We find that the model learns a simple binary-weighted dense circuit, i.e. the layer weights only take on two different values. This circuit effectively computes then stores all $m \choose 2$ outputs in superposition with some degree of noise, which can then be linearly read out with an additional linear layer. This circuit is only used when the inputs are sufficiently sparse. The same circuit design is used for almost all values of $d$, just with higher and higher noise from unrelated inputs.

The circuit design is fairly robust and general. It can be extended to other Boolean circuit operations straightforwardly. We supply a theoretical analysis for how this circuit works and contrast its efficiency to the sparse construction described in \citet{hanni2024}.


This circuit is particularly interesting as every intermediate neuron gives a useful contribution to every output. The model uses the increasing values of $d$ as opportunities to distribute each computation as widely as possible, reducing error. It does not form distinct, sparse, non-overlapping circuits, even for $d = {m \choose 2}$, where a naive perfect solution assigning each AND operation its own neuron would be possible.

Our contributions include:
\begin{itemize}
\item A novel construction for solving the Universal-AND problem with a 1-layer MLP with linear readout (\cref{binary_weighted_circuit}). We explain how the construction works and can be extended to other Boolean circuitry in one layer (\cref{circuit_analysis}).
\item Evidence that this construction can be learned in standard training dynamics (\cref{results}).
\item An approximate analysis of the asymptotic error of this circuit, with each input having variance $ \mathcal{O}(s^2 /d)$. (\cref{circuit_efficiency}).
\end{itemize}

\section{Related Work}



\citet[\textit{On the Complexity of Neural Computation in Superposition}]{comp_in_sup_complexity} establishes lower/upper parameter and neuron bounds for circuits such as Universal-AND.
Like \citep{hanni2024}, it supplies a sparse construction, and computes error bounds asymptotically. It also supplies an information-theoretic lower bound on the bits of parameters. Our model does not approach these theoretical limits, as we generously allocate parameters and only seek to restrict neurons. The paper establishes an exponential gap between the number of features that can be stored in an activation, and the number of computations that can be done in an equally sized network, demonstrating that models are likely to face strong pressures to compress computation to as few neurons as possible.

\citet{elhage2022toy} and \citet{scherlis2022polysemanticity} explore superposition in toy models and investigate how computation is done. The problem setups used in both cases are focused on representation and have largely trivial computation. They rely on a hidden layer smaller than the input size, which means they cannot easily distinguish between the computation in superposition and compressed computation.

\citet[\textit{Interpretability in Parameter Space: Minimizing Mechanistic Description Length with Attribution-based Parameter Decomposition}]{apd} also constructs toy models designed to be bottlenecked by a narrow hidden dimension, and introduces the conceptual difference between computation in superposition and compressed computation. They analyze the circuits through the Attribution Parameter Decomposition proposed in the paper and do not get an analytic description like that we derive in \cref{circuit_analysis}. Possibly the tasks chosen are too simple for interesting circuits to form.

\citet[\textit{Circuits in Superposition: Compressing many small neural networks}]{comp_in_sup_many} examines computation in superposition for a different non-trivial problem. They also focus on sparse theoretical constructions and their asymptotic behavior, and do not explore what models actually learn or if a dense circuit could perform better.

\section{Background And Setup}
\label{background}


\citet[\textit{Toward A Mathematical Framework for Computation in Superposition}]{hanni2024} first posed the question of how computation in superposition works. They introduce the \textbf{Universal-AND} problem, which computes the full set of pairwise AND operations on a set of inputs.

Formally, the Universal-AND problem considers a set of Boolean inputs \(v_1, \cdots, v_m\) taking values in \(\left\{0, 1\right\}\). The inputs are
\(s\)-sparse, i.e. at most \(s\) are active at once, \(\Sigma v_i \le s\). The problem is to create a one-layer MLP model that computes a vector of size \(d\), called the neuron activations, such that it's possible to read out every \(v_i \wedge v_j\) using an appropriate linear transform.

The problem as originally stated encodes the $m$ inputs in superposition in activation vector of size \(d_0\) but we omit this step and work directly on $v_i$.

For values of \(d \ll {m \choose 2}\) there is not enough compute to compute every possible AND pair separately. But if we assume \(s \ll d\), then the model can take advantage of the sparsity of the inputs to re-use the same model weights for unrelated calculations.

We model this as:

\[\mathbf{y} = \operatorname{ReLU}(W\mathbf{v}+\mathbf{b})\]

\[\mathbf{z}=R\mathbf{y} + \mathbf{c}\]

where \(W \in \mathbb{R}^{m\times d}, b \in \mathbb{R}^d, R\in \mathbb{R}^{d\times m^2}, c\in \mathbb{R}^{m^2}\).

In other words, \(W\) and \(\mathbf{b}\) describe the "compute layer", while \(R\) and \(\mathbf{c}\) describe the "readout layer". The existence of a readout layer makes the model effectively 2 layers deep. The readout weight matrix is large so the model is not bottlenecked on parameter count (unlike \citep{comp_in_sup_complexity}). Rather, the bottleneck is on the non-linear activations, the neurons. The readout layer should be understood as a trainable proxy for some more realistic setting, such as trying to learn a specific linear probe, or further model layers that are interested in a large fraction of possible pairwise AND operations.

For convenience of indexing, the output has dimension \(m^2\) corresponding to ordered pairs. It is symmetric, and the diagonal is unused.

\[\hat{z}_{im+j}= \left\{ \begin{aligned} v_i\wedge v_j & \quad& i \neq j \\ 0 & \quad & i=j \end{aligned} \right.\]

\section{Method}

We trained the above two-layer model on synthetic data where exactly \(s\) values of \(v_i\) are randomly chosen to be active (i.e. take value \(1\)).

We used RMS loss with different weighting per sample. This is because \(v_i\) are sampled uniformly, so test cases of the form \(0 \wedge 0\) are much more common than \(1 \wedge 0\) and \(1 \wedge 1\). So we up-weighted the loss so that the expected contribution from each of those three cases is equal. This encourages the model to focus on the few active results rather than the vast sea of inactive results. We can justify this because the second layer is intended as a "readout" layer. It is a proxy for a larger network that needs a lot of Boolean operations, which would also have unbalanced optimization pressure.

Some weight decay is used to regularize the network. This matches real-world training runs and encourages the model to focus on optimal circuits. Each model was trained with 6000 epochs of 10k batches to encourage this.

We used \(m=100, d=1000, s=3\) except where noted differently.

All experiments were trained on a single node with an inexpensive GPU. The full code can be found in the supplemental materials.

\section{Results}
\label{results}

\begin{figure}[htbp]
  \centering
  \begin{subfigure}[b]{0.3\textwidth}
    \includegraphics[width=\linewidth]{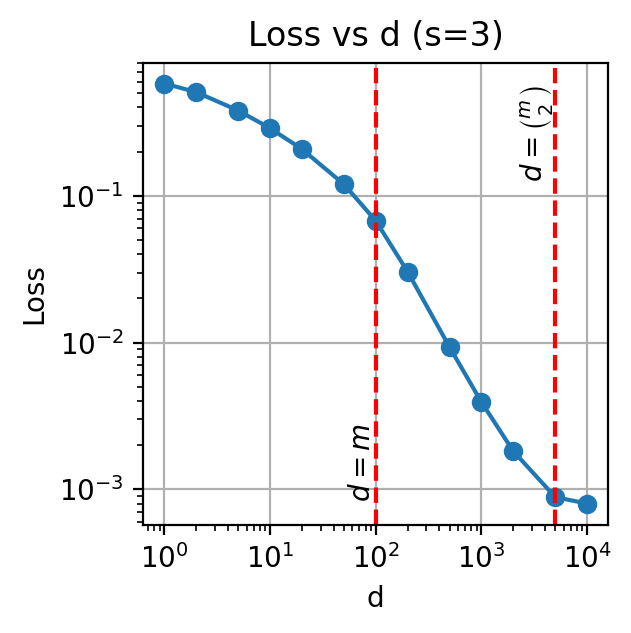}
    \label{fig:loss_v_d_s3}
  \end{subfigure}
  \hfill
  \begin{subfigure}[b]{0.3\textwidth}
    \includegraphics[width=\linewidth]{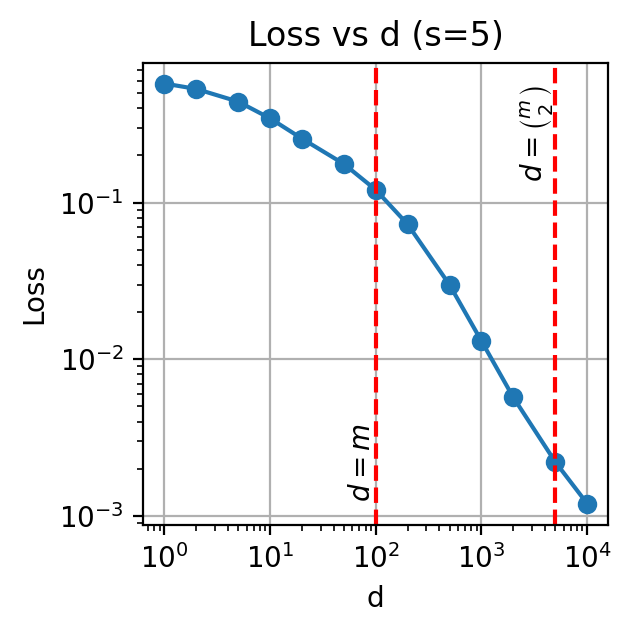}
    \label{fig:loss_v_d_s5}
  \end{subfigure}
  \hfill
  \begin{subfigure}[b]{0.3\textwidth}
    \includegraphics[width=\linewidth]{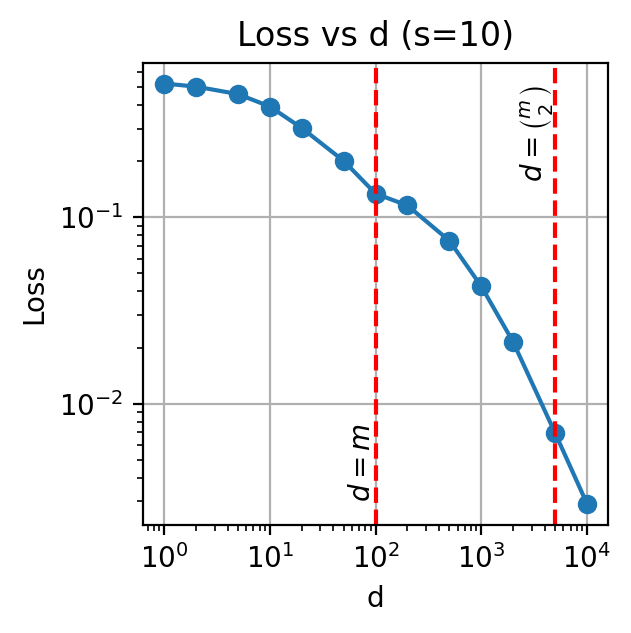}
    \label{fig:loss_v_d_s10}
  \end{subfigure}
  \caption{Model loss as $d$ increases.}
  \label{fig:loss_v_d}
\end{figure}

We find that at low \(s\) values, the model does find solutions that are capable of solving the Universal-AND problem, even extending to extremely low values of \(d\). The model weights take on a simple pattern of binary weights described below.

At higher values of \(s\) (starting at 10 for \(m=100\)), the model starts to prefer more degenerate solutions, particularly for \(d\leq m\). It either learns pure-additive circuits that roughly correspond to \(z_{im+j}=0.4(v_i+v_j)\) or it fails to update $W_{ij}$ from the initial values, and relies entirely on readout weights (discussed in \cref{just_feature_superposition}).

\subsection{The Binary Weighted Circuit}
\label{binary_weighted_circuit}

The model generally tends towards neuron weights that are binary, i.e. takes on only one of two different values. The choice seems randomized, with a roughly even balance between them.

\begin{figure}[htbp]
  \centering
  \includegraphics[width=0.8\textwidth]{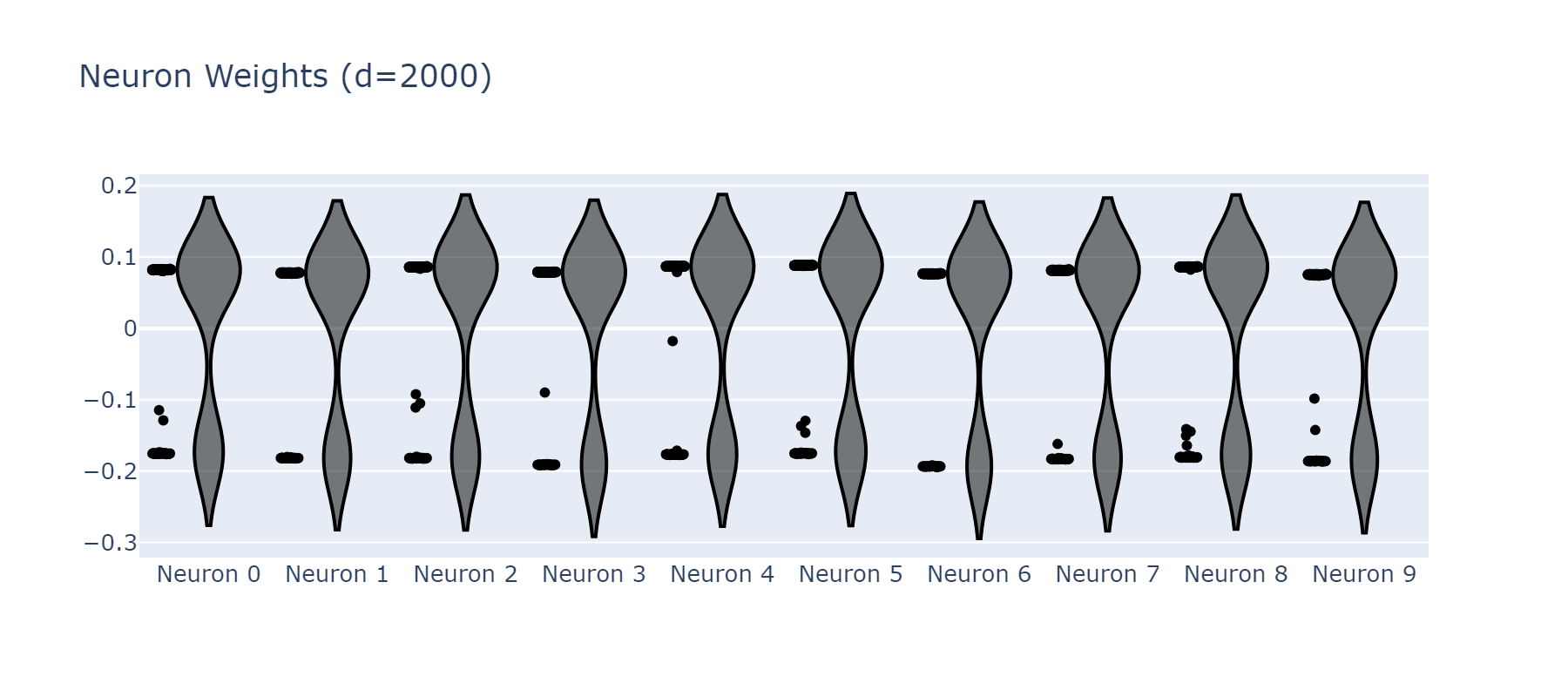}
  \caption{Distribution of $W_{ij}$ values for the first 10 neurons.}
  \label{fig:neuron_weight_violin}
\end{figure}

Expressed mathematically

\[W_{ij}=\left\{ \begin{aligned} u_i & \quad & \text{with probability }p_i\\ l_i & \quad & \text{with probability }1-p_i \end{aligned} \right.\]

We discuss why this is an effective choice in \cref{circuit_analysis}.

\begin{figure}[htbp]
  \centering
  \includegraphics[width=0.8\textwidth]{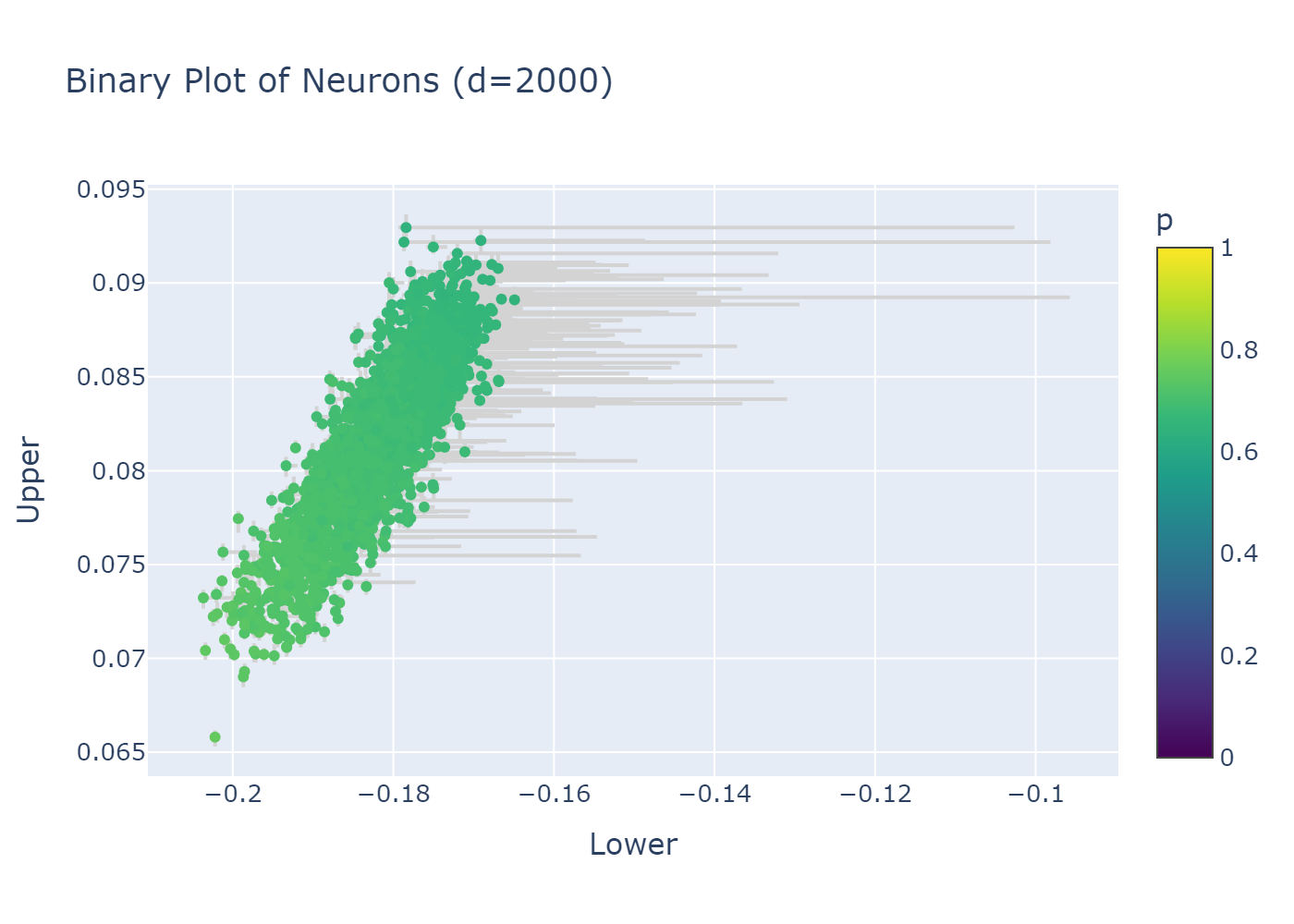}
  \caption{Distribution of upper/lower weights by neuron. Error bars show 90th-percentile deviation from the weight being near to either the upper/lower bound.}
  \label{fig:binary_plot_neurons}
\end{figure}

We chart the specific values of \(u_i\), \(l_i\), \(p_i\) in shown in \cref{fig:binary_plot_neurons}. It illustrates that all neuron weights are clustered in a tight region and $u_ip_i + l_i(1-p_i)\approx 0$.

Charts for more values of $d$ can be found in \cref{binary_weight_charts_appendix}.

\subsection{Readout Charts}
\label{readout_charts}

Another way of viewing the neurons is in terms of how they are read out by matrix \(R\). Pick two arbitrary inputs (say \(v_1\) and \(v_2\)), then plot each neuron in a scatter chart based on their weights (i.e. $W_{i1}$ on the x-axis, $W_{i2}$ on the y-axis). Color the neurons based on their readout weight for \(v_1 \wedge v_2\) (i.e. \(R_{(1m+2),i}\)).

\begin{figure}[htbp]
  \centering
  \begin{minipage}[b]{0.45\textwidth}
    \centering
    \includegraphics[width=\textwidth]{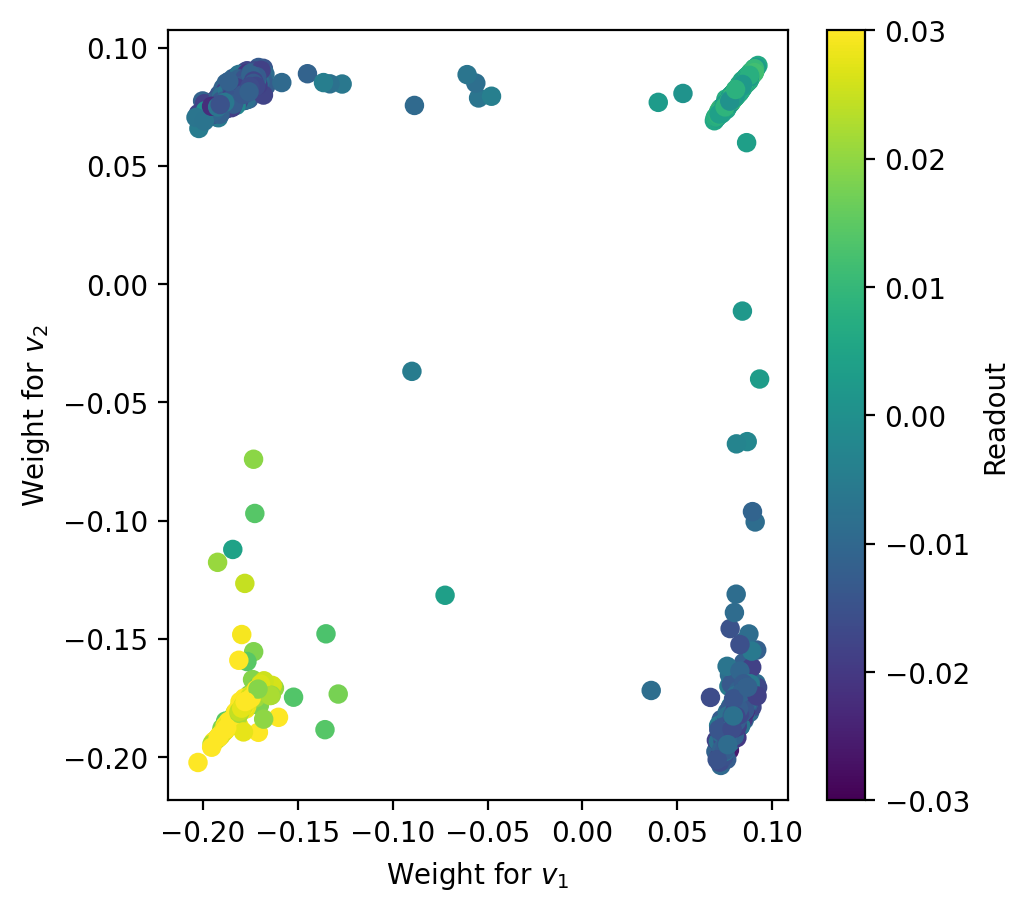}
    \caption{Readout weights by input weights ($d=2000)$}
    \label{fig:readout_d2000}
  \end{minipage}
  \hfill
  \begin{minipage}[b]{0.45\textwidth}
    \centering
    \includegraphics[width=\textwidth]{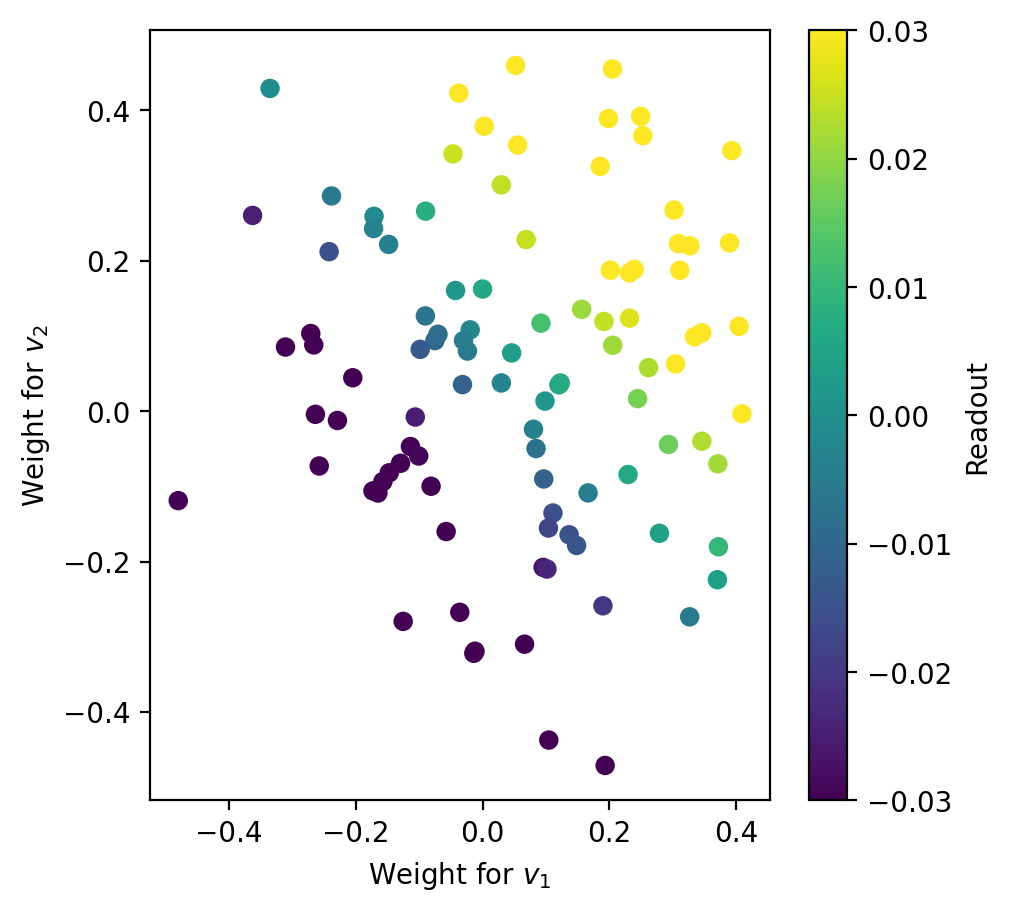}
    \caption{Readout weights by input weights ($d=100, s=50$)}
    \label{fig:readout_d100_s50}
  \end{minipage}
\end{figure}

The four corners of \cref{fig:readout_d2000} correspond to classes A (top right), C (bottom left), B1, and B2 as described in \cref{circuit_analysis}. Class C has higher weights per-neuron as there are fewer neurons in that class than class A.

Even in cases where the weights don't form a Binary Weighted Circuit, these charts give a clear indication of how readout works. \Cref{fig:readout_d100_s50} shows $s=50$, where there is not enough sparsity for the Binary Weighted Circuit. The circuit that forms instead is easily distinguished by the lack of class C neurons, and more uniform distribution of weights.

\section{Analysis}
\label{analysis}
\subsection{Circuit Analysis}
\label{circuit_analysis}

The results show that for a wide range of parameters, the model tends towards binary weights where for any given neuron, the weights it uses only take on two different values, with no discernal pattern. Expressed mathematically

\[W_{ij}=\left\{ \begin{aligned} u_i & \quad & \text{with probability }p_i\\ l_i & \quad & \text{with probability }1-p_i \end{aligned} \right.\]

This resembles constructions from \citep{hanni2024}. Their main result\footnote{\citet{hanni2024} also mentions a randomized dense construction which is also similar. See \cref{just_feature_superposition}} is a sparse construction which followed this exact pattern. We'll call it the \textbf{CiS Construction}. It takes on values

\[u_i = 1\quad l_i=0\quad p_i= \log^2m/\sqrt{d}\]

In the CiS Construction \(p_i\) is small, and \(l_i\) is zero, meaning the neurons were only sparsely connected. This property was key in proving an upper bound on the loss of the model.

But in the learned Binary Weighted Circuit we see quite different values. These values are representative\footnote{As the readout matrix \(R\) can supply an arbitrary positive scaling, the only important details of \(u\) / \(l\) are their signs and ratio. The magnitude tends to be around \(1/\sqrt{d}\) as this minimizes regularization loss (weight decay) in a two layer network. \Cref{binary_weight_charts_appendix} includes more details on the exact values seen.}:

\[u_i = 0.1\quad l_i=-0.25\quad p_i=0.75\]

\textit{Notation: When the values are a constant, we'll omit the \(i\) subscript.}

Unlike the CiS Construction, this is dense. Every neuron reads a significant value for every possible input.

We now explain how neurons in this architecture can be used to approximate the AND operation with a linear readout. Without loss of generality, we focus on the first two input variables—that is, we aim to read out \(v_1 \wedge v_2\) from the activations of $d$ neurons that share the same values for $u$, $l$, and $p$, differing only in their randomly initialized weights.

We can subdivide the neurons into 4 classes based on the weight used for each of the first two inputs.

\[\begin{aligned} 
\text{Class A} & \quad & y_i = \operatorname{ReLU}(&uv_1+uv_2+X_i+b) &\quad & \text{occurs with proportion }p^2\\
\text{Class B1} & \quad & y_i = \operatorname{ReLU}(&uv_1+lv_2+X_i+b) &\quad & \text{occurs with proportion }p(1-p)\\ 
\text{Class B2} & \quad & y_i = \operatorname{ReLU}(&lv_1+uv_2+X_i+b) &\quad & \text{occurs with proportion }p(1-p)\\
 \text{Class C} & \quad & y_i = \operatorname{ReLU}(&lv_1+lv_2+X_i+b) &\quad & \text{occurs with proportion }(1-p)^2\\ \end{aligned}\]

For simplicity of analysis, we'll ignore contributions from other inputs ($X_i$, the interference term). It can be treated like a mean zero random variable of noise. \Cref{circuit_efficiency} looks at $X_i$ closer. We'll also set \(b=0.05\); the exact choice of constants is not relevant to the key argument.

\Cref{tab:truth_table} shows a truth table for the results of each class of neuron for the 4 possible values of \(v_0,v_1\).

\begin{table}[h!]
\centering
\begin{tabular}{cc|cccc|c}
\toprule
$v_1$ & $v_2$ & A & B1 & B2 & C & $4(A + C - B1 - B2)$ \\
\midrule
0 & 0 & 0.05 & 0.05 & 0.05 & 0.05 & 0 \\
0 & 1 & 0.15 & 0    & 0.15 & 0    & 0 \\
1 & 0 & 0.15 & 0.15 & 0    & 0    & 0 \\
1 & 1 & 0.25 & 0    & 0    & 0    & 1 \\
\bottomrule
\end{tabular}
\caption{Approximate truth tables for each neuron class, and their linear combination.}
\label{tab:truth_table}
\end{table}

Taking the right linear combination of the 4 truth tables, we can recreate the AND truth table. This linear combination is a close match for the values seen in \cref{readout_charts}.

Taking a linear combination will always be possible if the 4 classes are linearly independent, which is true for quite a wide range of choices for the key parameters.

Of course, the interference term $X_i$ means that these truth tables are not accurate. But in the problem setup there is no correlation between inputs (beyond the sparsity constraint), the different inputs are almost completely independent. So the interference term can be modeled as noise. The noise will pull the 4 classes towards co-linearity, but for an appropriate bias value there will always be a difference that can be exploited.

As there are many neurons in each class, the readout matrix can average over them, reducing the noise to reasonable levels. This is similar to the proofs on noise bounds in \citet{hanni2024}, discussed further in \cref{circuit_efficiency}.

Finally, we note that the same argument and weights matrix can be adapted for other tasks. A different choice linear combination in the readout matrix can supply any other truth table instead (\cref{xor_circuits}). And multiple inputs can be considered: with 3 inputs there are 8 possible neuron classes, to form in linear combination the 8 values of a 3-way truth table\footnote{Error grows swiftly with number of inputs, so more complex circuits in models no doubt rely on multiple layers, discussed in \citep{hanni2024}.}.

\subsection{Circuit Efficiency}
\label{circuit_efficiency}

Why is this dense Binary Weighted Circuit learned in preference to the CiS Construction described in \citet{hanni2024}? We present an approximating argument that it produces lower loss values.

To recap, the CiS Construction fills the weight matrix with zeros and ones, with the ones being at probability \(p\ll 1\). While the Binary Weighted Circuit fills the weight matrix with \(u\) and \(l\), with \(l < 0 <u\) and \(p\approx0.5\).

Recall the neuron-oriented view from the previous section where we have two distinguished inputs, w.l.o.g. $v_1$, $v_2$:

\[y_i =\operatorname{ReLU}(lv_1+uv_2+X_i+b)\]

The contribution from all other inputs is folded into an interference term \(X_i\)

\[X_i=\sum_{j=3}^m W_{ij}v_j\]

Most of the entries of \(X\) don't contribute, as \(v_j\) will be zero. There will be at most \(s\) entries, each randomly contributing \(u\) or \(l\) depending on \(W_{ij}\). So we can approximate \(X\) as binomially distributed\footnote{This approximation elides the correlation between \(X\) and \(v_1\), \(v_2\), which for high values of \(p\) can be important, but doesn't effect the circuits examined significantly.}.

\[X_i \sim (u-l)\operatorname{Binom}(s, p)+sl\]

\[\operatorname{Var}(X_i)=(u-l)^2sp(1-p)\]

We can treat the collection of \(X_i\) as independent, as the randomness comes from independent choices of \(W_{ij}\).

The variance of $\operatorname{ReLU}(X)$ can't be exactly characterized. We'll approximate it scales variance by a constant factor, \(\beta\), giving:

\[\operatorname{Var}(y_i)=\beta(u-l)^2sp(1-p)\]

Now we can use this estimate to compute the variance of the readout values.

In the CiS Construction, the readout is the mean value of all neurons that are in class A (i.e. have a weight of 1 for \(v_1\) and \(v_2\)). There are \(dp^2\) such neurons in expectation so we can divide by that to get the variance of the mean.

\[\operatorname{Var}_\text{CiS}(z_i)=\beta s p (1-p) /dp^2\]

Meanwhile in the Binary Weighted Circuit, we pick readout weights based on the inverse of the \(4\times 4\)  truth table matrix. It can be shown that the total weight for each class is approximately \(\pm4\sqrt{s}\)
\footnote{This comes from \(X_i\) having variance proportional to \(s\), but the neuron classes only differ from each other by a constant translation of at most \(2(u-l)\). As \(s\) increases, the classes ReLU zero-points are at increasingly similar points on the probability distribution.}.

So we have \(dp^2\) class A neurons, each with readout weight  \(\pm4\sqrt{s}/dp^2\). This gives total variance

\[\operatorname{Var}_\text{Binary}(z_i)=\beta(0.1--0.25)^2sp(1-p) \left( dp^2\left(\frac{4\sqrt{s}}{dp^2}\right)^2+... \right)\]

\[=0.35^2 4^2 s \beta sp(1-p) \left( \frac{1}{p^2}+\frac{2}{p(1-p)}+\frac{1}{(1-p)^2} \right)/d\]

Both variance formulas are pretty similar in terms of constants and asymptotic behavior, which is unsurprising as so far we've just evaluated different choices of \(u\) and \(l\). The dense case is worse by a factor of \(s\), and includes some extra terms that were previously ignored because \(p\) is small in the CiS Construction.

To compare these properly, we need to consider the different values of \(p\). For the CiS Construction, the value of $p$ was asymptotically small, \(\log^2m/\sqrt{d}\), while in the Binary Weighted Circuit \(p\) is a constant \(\approx0.75\).

Substituting and simplifying, we get

\[\operatorname{Var}_\text{CiS}(z_i)=\mathcal{O}(s /dp )=\mathcal{O}(s / \sqrt{d} / \log^2 m)\]

\[\operatorname{Var}_\text{Binary}(z_i)= \mathcal{O}(s^2 /d)\]

So the Binary Weighted Circuit has superior efficiency when \(s\) grows slower than \(\sqrt{d}/\log^2m\).

This result matches intuition. Increasing \(p\) makes the circuitry more dense, i.e. the model is making use more neurons for each calculation. This increases the variance of individual neurons but gives you many more to average over. \(s\) determines that per-neuron noise, so determines the trade-off. The CiS Construction merely aimed to minimize the interference term of a single neuron, as this was critical for establishing provable error bounds.

Aside from accuracy considerations, dense circuits are more likely to result from training than equivalent sparse ones, as they score better for regularization loss. Adding weight decay adds an inductive bias towards "spreading out" circuits as much as possible.

\section{Limitations}

This paper attempts to build on previous theoretical understanding in a more realistic trained setting, but toy models still fall short of real-world models. In particular, our use of monosemantic inputs/outputs and choice of the Universal-AND problem are deliberate simplifications of superposition, and complex circuits found in practice.

The experimental results show that the Binary Weighted Circuit is used at reasonable values of sparsity (\cref{fig:binary_distribution}), but we have not performed a full sweep to fully characterize this. Nor have we explained why the particular values of $u$, $v$, $p$ observed are used. \citet{comp_in_sup_complexity} gives much tighter and general bounds on what is possible, so we have focused on the mechanics of the circuitry.

\Cref{analysis} relies on several approximations that are not rigorously proved. Future theoretical or empirical work would be needed to gain confidence in these claims.

\section{Conclusion}
 



We have established the dense Binary Weighted Circuit that solves the Universal-AND problem. We analytically describe the fundamental behaviour of the circuit and approximate its error rate. This represents a useful formulation for understanding compressed computation - previous works either described theoretical sparse circuits that are not used in practice, or do not give an analytic description of the circuit. By using an explicitly monosemantic input/output, it helps answer the question posed in \citet{apd}: whether compressed computation and computation in superposition are \say{subtly distinct phenomena}.

Dense circuits like these challenge the common assumption that circuits can be found by finding a sparse subset of connections inside a larger model, and give an additional explanation why features are rarely monosemantic. If it is better to have many shared noisy calculations than a smaller set of isolated, reliable ones, then a different set of techniques is needed to detect them. It is an important principle to be aware of while conducting interpretability work, or designing new network architectures.

Given the novel structure of this circuit combined with its simplicity, we believe this problem and circuit can act as a good testbed for circuit-based interpretability tooling.

\newpage

\bibliographystyle{plainnat}
\bibliography{sample}

\newpage

\appendix

\section{Other Considerations}

We include brief notes on how our result interacts with related discussions regarding Computation In Superposition.

\subsection{Is this Just Feature Superposition?}

\label{just_feature_superposition}
In a sense, the circuit found here bears a lot of resemblance to simply randomly embedding the \(m\) inputs in \(d\) dimensional space. In both cases, you get a mix of neurons with different response patterns, and you can approximate the AND operation by taking a dense linear combination of the neurons. 

\begin{figure}[htbp]
  \centering
  \includegraphics[width=0.8\textwidth]{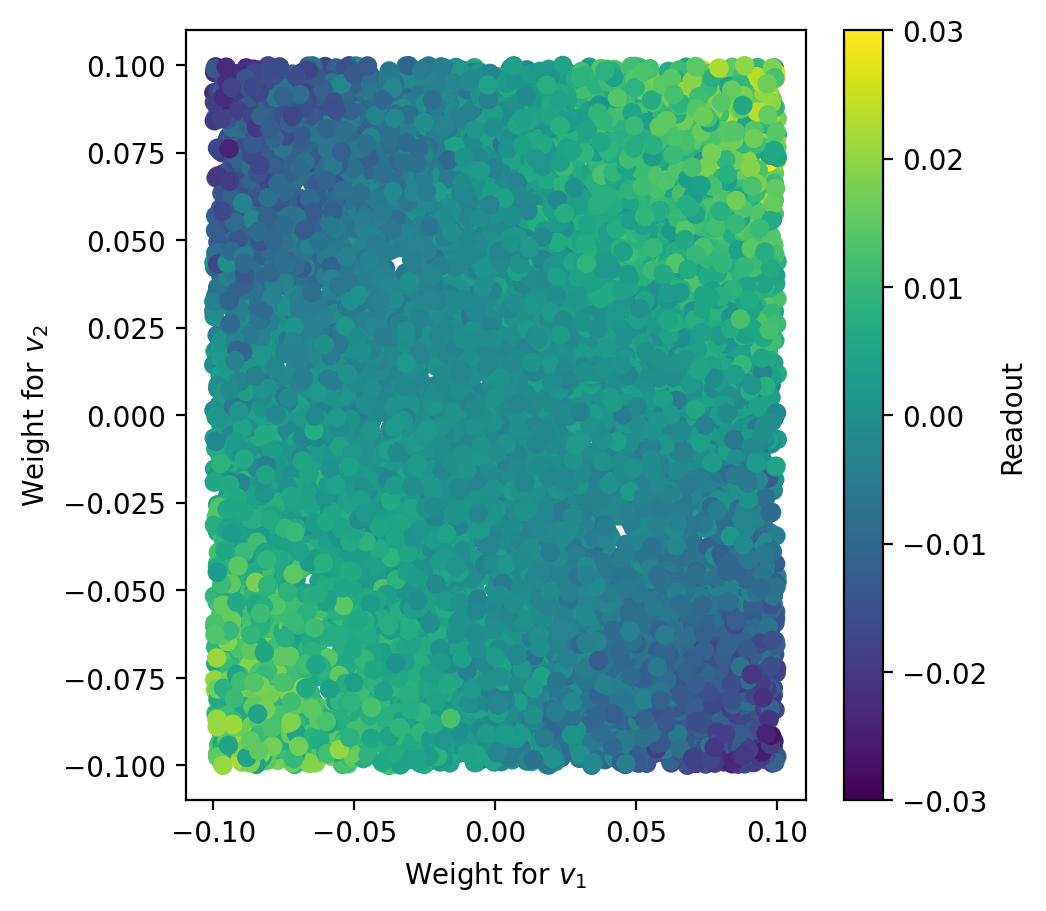}
  \caption{Readout weight by input weights (uniform initialization, $d=10000$)}
  \label{fig:readout_frozen}
\end{figure}

Indeed, training the toy model with the first layer frozen to random values still results in a similar pattern of readouts (\cref{fig:readout_frozen}). \citet{hanni2024} includes a similar observation in Section 3.3.

\begin{figure}[htbp]
  \centering
  \includegraphics[width=0.8\textwidth]{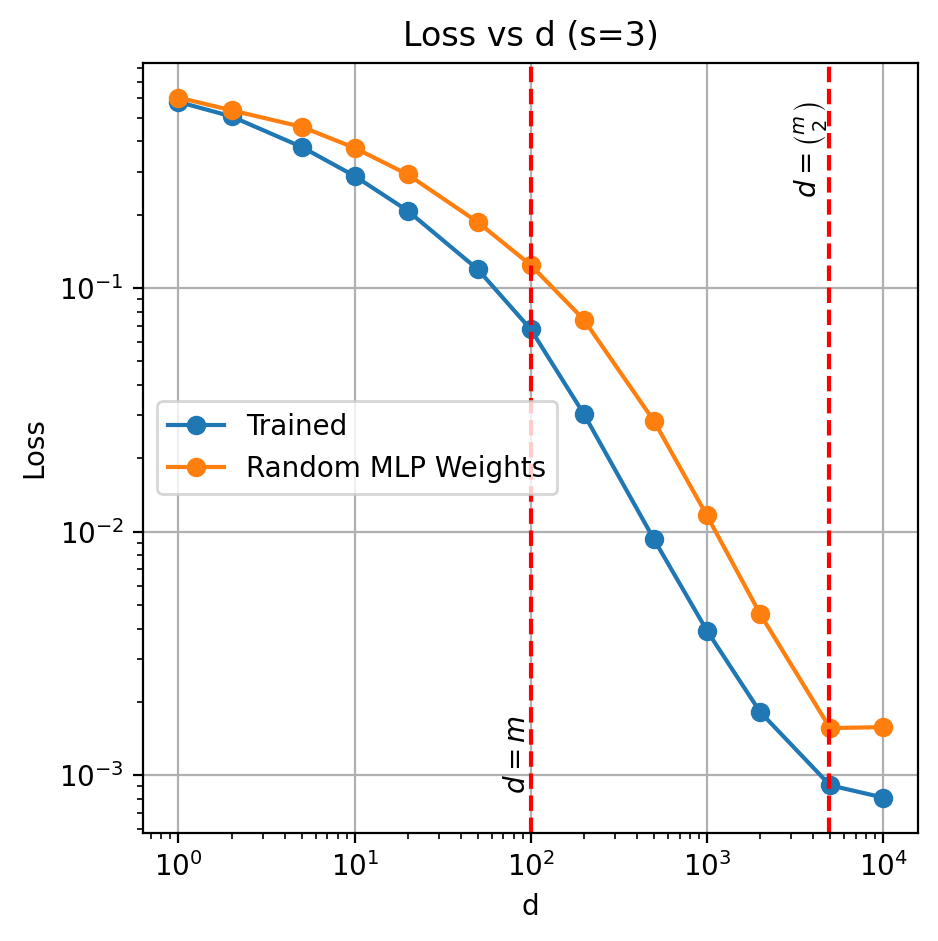}
  \caption{}
  \label{fig:loss_v_d_randomized}
\end{figure}

The circuit described in this paper uses binary weights, rather than some other random distribution. But we expect such a pure distribution is unlikely to replicate outside of toy models. Using binary weights simplifies analysis and has an improved loss, but only improves loss by a constant factor (\cref{fig:loss_v_d_randomized}). 

Thus an alternative reading of this paper's results would be: For a set of sparse binary features stored in superposition with sufficiently random directions, it is possible to linearly readout any Boolean circuit, subject to a certain amount of error.

\subsection{XOR Circuits}

\label{xor_circuits}

\cite{xors_repr} observes that readout directions corresponding to XOR of input features commonly occur in models. Marks argues that these will cause linear probes to fail to generalize. I.e. if a probe trained to predict \(v_1\) only on data where \(v_2=0\), then it will be just as likely to identify \(v_1 \oplus v_2\) as it is \(v_1\). These give opposite answers when out of distribution \(v_2=1\), breaking the probe.

Our result gives an explanation for why XOR circuits may be readaoutable, but probes still have some generalizability. Using the parameters described above, the learned readout weights corresponding to the aggregate output of each neuron class yield the following linear combinations:

\[v_1 \sim 4A-4B1+\frac{8}{3}B2-\frac{8}{3}C\]

\[v_1 \oplus v_2 \sim \frac{20}{3}B1+\frac{20}{3}B2-\frac{40}{3}C\]

While both target functions are linearly decodable, the XOR direction requires significantly larger weights. As a result, under typical regularization schemes\footnote{For some choices of \(p\) when using \(L_1\) norm relative preference may reverse. Nonetheless, the general principle holds: one of the two linearly accessible directions will be favoured.} the $v_1$ direction is likely to be favoured.

\section{Binary Weight Charts}
\label{binary_weight_charts_appendix}

We supply binary weight charts for a range of values of $d$. Recall that \(m=100, s=3\). Error bars show 90th-percentile, so indicate the extent to which the individual weights associated with a neuron do not perfectly match $u$ / $l$.

\begin{figure}[htbp]
  \centering
  \includegraphics[width=0.8\textwidth]{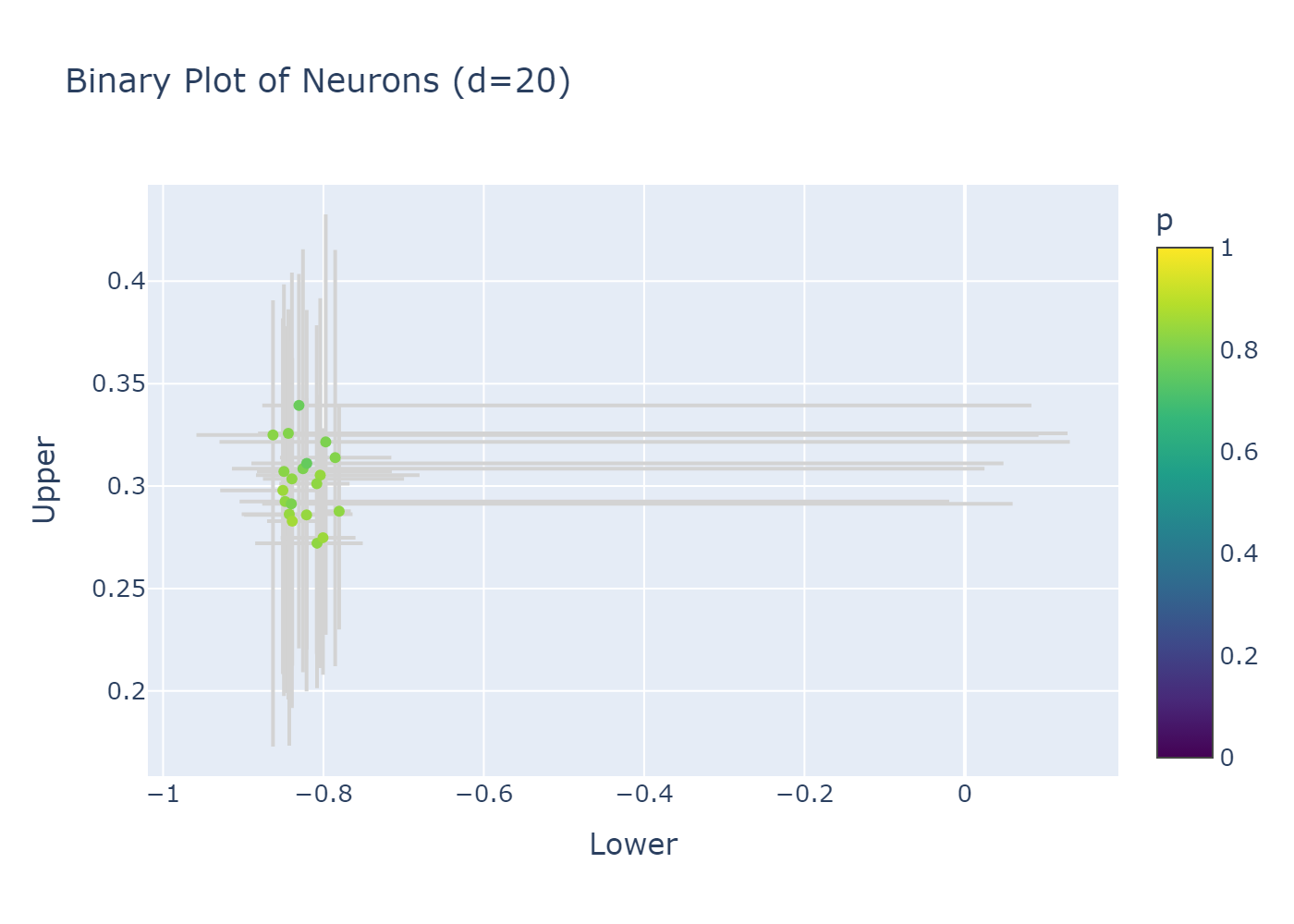}
\end{figure}

\begin{figure}[htbp]
  \centering
  \includegraphics[width=0.8\textwidth]{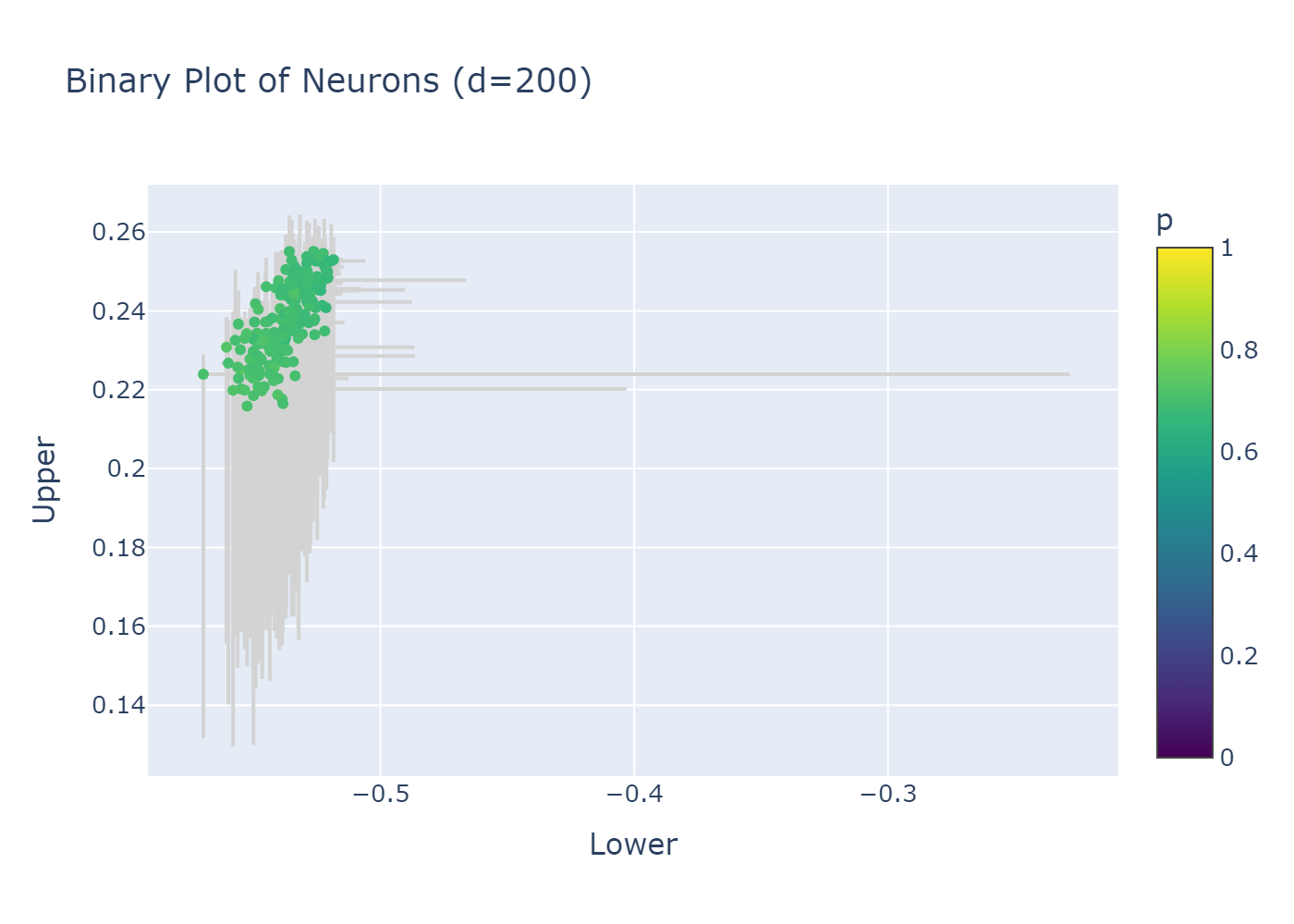}
\end{figure}

\begin{figure}[htbp]
  \centering
  \includegraphics[width=0.8\textwidth]{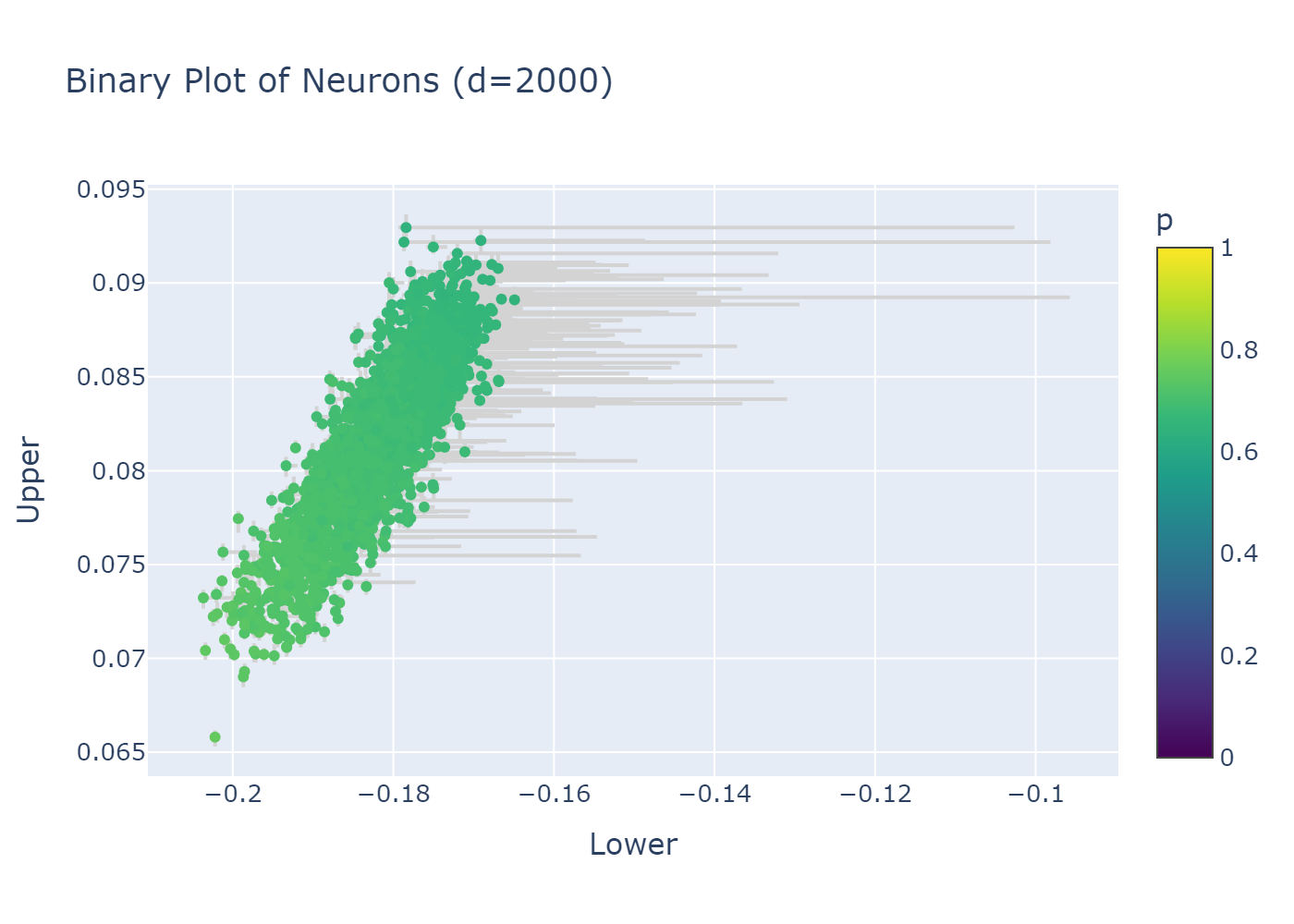}
\end{figure}

\begin{figure}[htbp]
  \centering
  \includegraphics[width=0.8\textwidth]{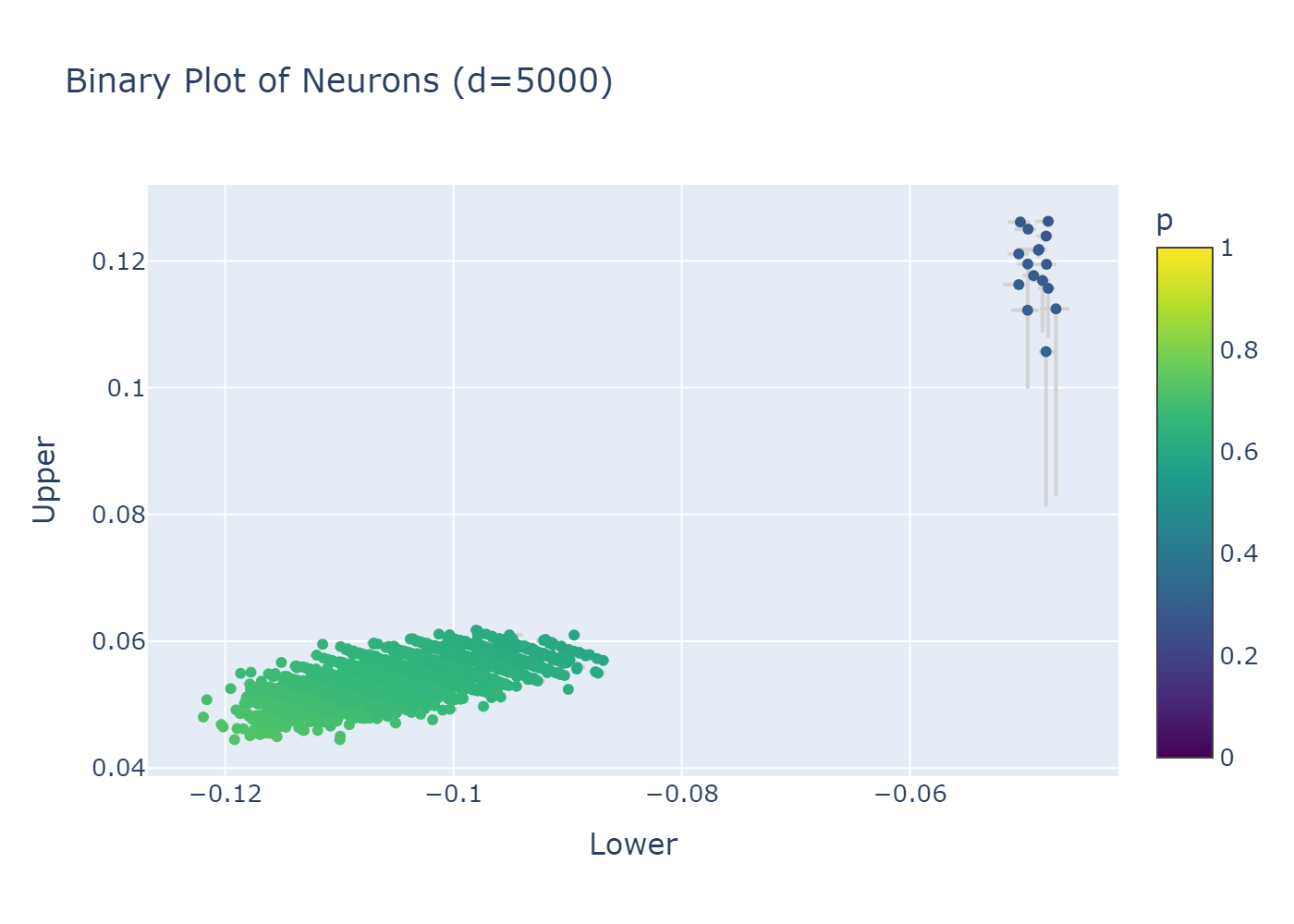}
\end{figure}

\begin{figure}[htbp]
  \centering
  \includegraphics[width=0.8\textwidth]{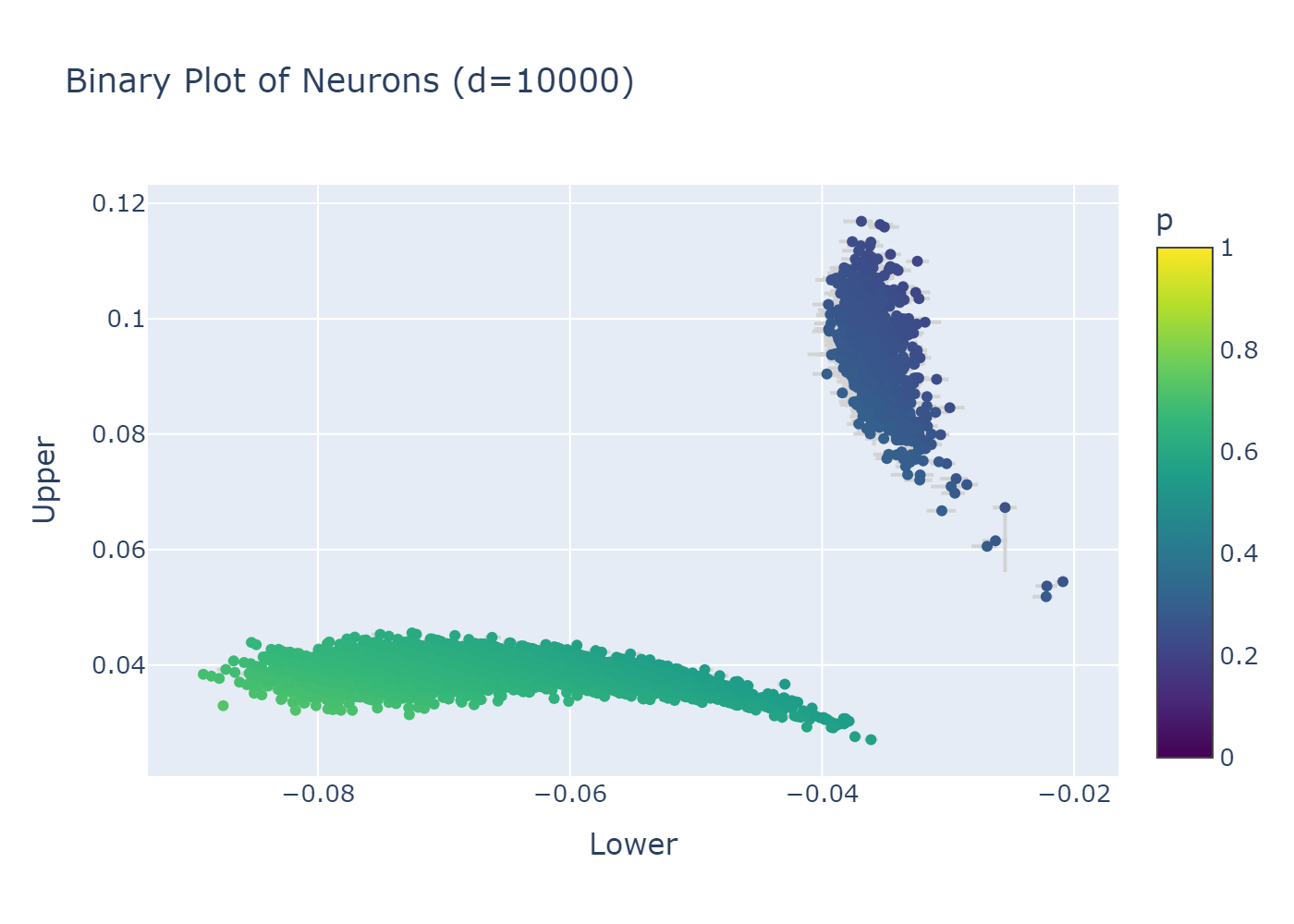}
\end{figure}

\begin{figure}[htbp]
  \centering
  \includegraphics[width=0.8\textwidth]{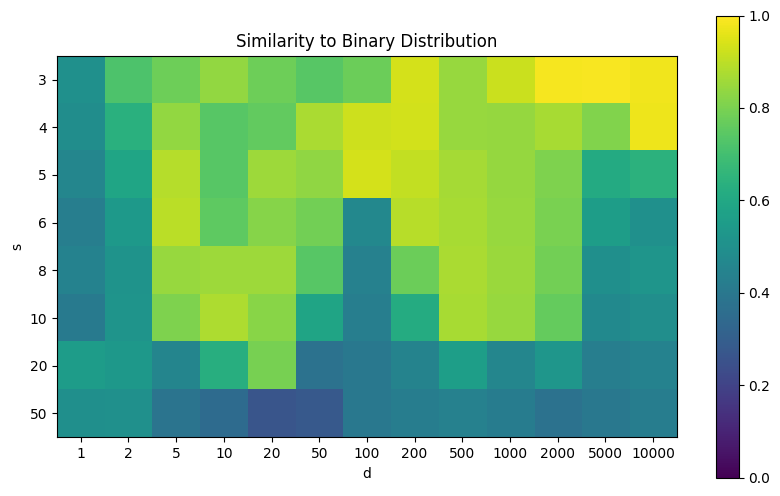}
  \caption{Similarity of learned weight matrix to a binary distribution for various $d$, $s$}
  \label{fig:binary_distribution}
\end{figure}

In \cref{fig:binary_distribution} we measure the similarity of the entire weights matrix to a binary distribution, using 

\[ \text{score} = 1-\underset{i, j}{\text{mean}} \frac{\min(|W_{ij}-u_i|,|W_{ij}-l_i|)}{2(u_i-l_i)}\quad\text{ for }u_i = \max_j W_{ij}, l_i = \min_j W_{ij}\]. 

\end{document}